\documentclass[journal, 12pt,onecolumn, draftclsnofoot]{ieeeconf}      
\IEEEoverridecommandlockouts                              
\usepackage{epsf,graphicx}
\usepackage{latexsym,amssymb}
\usepackage{setspace,cite}
\usepackage{amsmath}
\usepackage{float}
\usepackage{subfloat}
\usepackage{anysize}
\usepackage{tabularx}
\usepackage{subcaption}
\usepackage{hyperref}
\usepackage[affil-it]{authblk}

\title{\LARGE \bf
Deep Convolutional Poses for Human Interaction Recognition in Monocular Videos
}

\author{Marcel Sheeny de Moraes\textsuperscript{1}, Sankha Mukherjee\textsuperscript{1}, Neil M Robertson\textsuperscript{1}}
\affil{\textsuperscript{1}School of Engineering \& Physical Sciences, Heriot-Watt University}

\begin{document}

\maketitle
\thispagestyle{empty}
\pagestyle{empty}

\begin{abstract}

Human interaction recognition is a challenging problem in computer vision and has been researched over the years due to its important applications. With the development of deep models for the human pose estimation problem, this work aims to verify the effectiveness of using the human pose in order to recognize the human interaction in monocular videos. This paper developed a method based on 5 steps: detect each person in the scene, track them, retrieve the human pose, extract features based on the pose and finally recognize the interaction using a classifier. The Two-Person interaction dataset \cite{yun2012two} was used for the development of this methodology. Using a whole sequence evaluation approach it achieved 87.56\% of average accuracy of all interaction. Yun, \textit{et at} \cite{yun2012two} achieved 91.10\% using the same dataset, however their methodology used the depth sensor to recognize the interaction. The methodology developed in this paper shows that an RGB camera can be as effective as depth cameras to recognize the interaction between two persons using the recent development of deep models to estimate the human pose.

\end{abstract}

\section{INTRODUCTION}

Human interaction recognition is a challenging problem in computer vision and has been researched over the years due to its important applications. The main applications are in the fields of surveillance, human-robot interaction and automatic video-labelling.

There are many approaches to recognize human action. One common approach is to use bag-of-words \cite{yang2007evaluating} where handcrafted features are used, relevant works which used this approach can be seen in \cite{vahdat2011discriminative, patron2010high, slimani2014human}. One key information to recognize the human action is the human pose estimation, it means to predict where is each part of the human body (arms, legs and head). The pose can be captured with high accuracy using a depth camera \cite{zhang2012microsoft, shotton2013real}, however using a monocular camera the results were not accurate until researchers started to use  deep convolutional neural networks in order to estimate the human pose.

\begin{figure}[t]
    \centering
    \includegraphics[width=0.4\paperwidth]{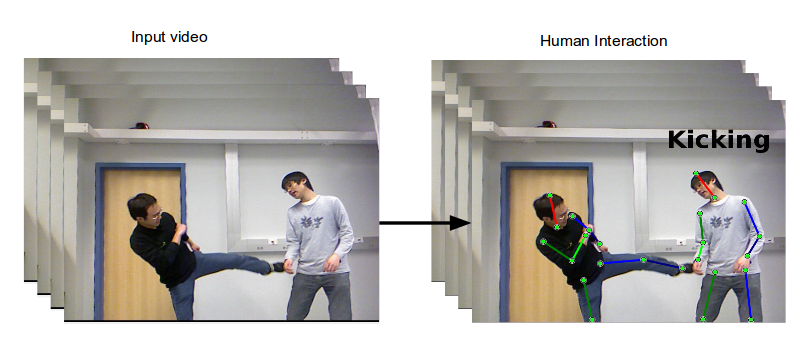}
    \caption{Example of interaction to be recognized, finding the pose estimation and classifying the interaction.}
    \label{fig:problem_example}
\end{figure}

For surveillance applications, most of the time only a monocular RGB video is provided to recognize the interaction. With the development of deep models for the human pose estimation problem, this paper has the intention to verify the effectiveness of using the pose in order to recognize the human interaction in monocular videos. The Figure \ref{fig:problem_example} visualizes the goal of this dissertation.

In order to recognize the human interaction, a method was developed based on the following 5 steps: detect each person in the scene, track them, retrieve the human pose, extract features from the pose information and finally recognize the interaction using a classifier.

The Two-Person interaction dataset \cite{yun2012two} was used, it contains 282 samples with 8 different types of interaction. Using a whole sequence evaluation approach the methodology developed achieved 87.56\% of average accuracy of all interaction using whole evaluation. Yun, \textit{et at} \cite{yun2012two} achieved 91.10\% using the same dataset, however they used the depth sensor to capture the human pose and this paper just uses RGB information to capture the pose.

The main contribution of this paper is that the recent development of human pose estimation could achieve close results to a method developed using a depth camera. The methodology presented in this document shows that retrieving the human pose from an RGB camera to recognize the interaction between two persons can be as effective as depth cameras.
   
\section{Related works}
\label{state_of_the_art}

\subsection{Human Pose Estimation}

Many papers started to use deep convolutional neural networks (DCNN) for human pose estimation (HPE) due to its effectiveness in many computer vision tasks \cite{toshev2014deeppose, tompson2015efficient, pishchulin2015deepcut}. The current state-of-the-art results in human pose estimation uses DCNN which was developed by Wei, \textit{et al}, 2016 \cite{wei2016convolutional}. It achieved 90.5\% of accuracy PCKh at 0.2.

\subsection{Human interaction recognition}

One of the first research papers in human interaction recognition was developed by Johnson, \textit{et al}, 1998 \cite{johnson1998acquisition}, which uses a dataset with two persons that segments each person in the scene and create a Spline type contour to have a silhouette. Then some points are samples from the silhouette and a Hidden Markov Model (HMM) is used to classify the interaction. This paper achieved 95\% of accuracy.

A paper published by Yun, \textit{et at,} 2012 using a two person dataset interaction, retrieves the human pose estimation from a Kinect device. From the pose estimation it retrieves the joint distances, joint motion, plane, normal place, velocity and normal velocity as features to be used in Multiple Instance Learning (MIL) classifier. It achieved 80.30\% of accuracy getting 3 frames of information and 91.1 \% using the whole sequence.

Using the same dataset as Yun, \textit{et at,} \cite{yun2012two}, Hu \textit{et at,} 2013, \cite{hu2013efficient} used the skeleton captured by the Kinect, got the information of the active person doing the action and computed the same features of Yun, \textit{et al} (joint distances, joint motion, plane, normal place, velocity and normal velocity). It used and Hidden Markov Model (HMM) to classify the interaction per frame and Multiple Instance Learning (MIL) to recognize the whole sequence. It achieved 76.1\% of accuracy for each frame and 83.33 \% of accuracy recognizing the whole sequence.

Zhu, \textit{et at} 2015 \cite{zhu2015co} creates a Deep LSTM network getting the skeleton data from the Two-Person Interaction dataset \cite{yun2012two}. It achieved 90.41\% of mean accuracy between all interactions.

\section{Methodology}

The main idea of this paper is to take the advantage of the development of deep learning in human pose estimation in order to solve the problem of human interaction. 

To use the human pose estimation to classify the interaction, first we need to detect the humans in the scene, since the human pose method used is not robust for multiple people. Then we need to track them, since the detection method is not perfect. After tracked we can get the human pose estimation to finally recognize the interaction.

\begin{figure*}[t!]
  \centering
      \includegraphics[width=1\textwidth,height=7cm]{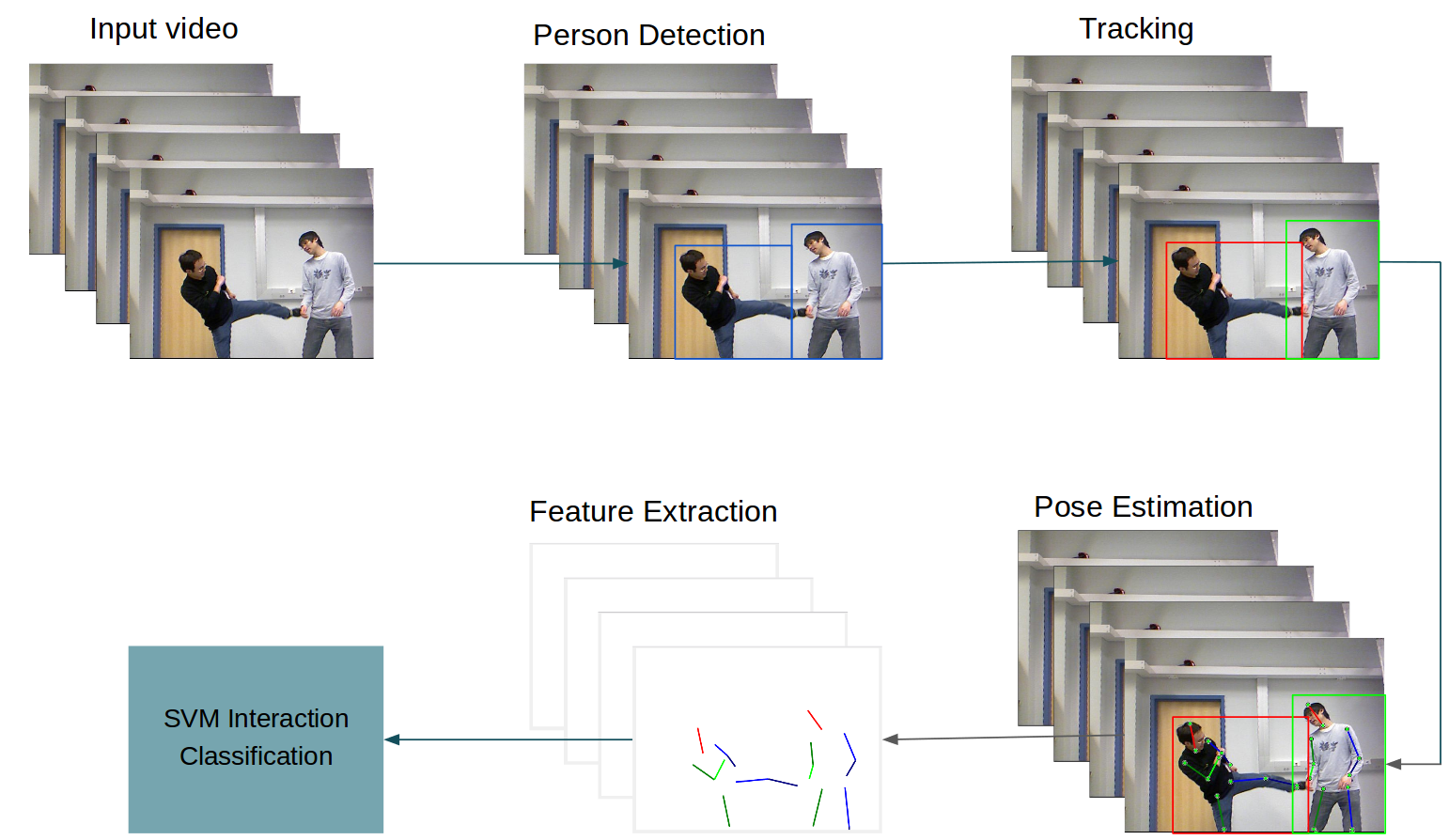}
  \caption{Overview of the methodology used for interaction recognition.}\label{fig:overview}
\end{figure*}

Similarly to the method developed by Park and Aggarwal, 2004 \cite{park2004hierarchical}, who detected and tracked each part of the human body, however it was done in a high-level way, it did not detect symmetric parts, it detects head, upper-body and low-body. The idea to solve this problem is to extract features similar to the ones used by Yun, \textit{et al.} \cite{yun2012two} where they get the human pose estimation captured from the depth sensor and classify the interaction using the joints information. The main difference between the work developed by Yun, \textit{et al.} is that the depth information is not being used, only RGB. The overview from the whole methodology can be visualized in the Figure \ref{fig:overview} which will be described in the following sections

\subsection{Person Detection}
\label{sec:person_detection}

Since the method to retrieve the human pose is not robust to multiple people, a person detector method is needed. The technique developed by Ren, \textit{et al.}, 2014, "Faster R-CNN: Towards real-time object detection with region proposal networks" \cite{ren2015faster} was used, it achieves state-of-the-art results for object detection using deep convolutional neural networks. This technique can detect the objects in 0.1 seconds.

The Faster R-CNN \cite{ren2015faster} is done in two steps, first it trains a network to generate bounding boxes, then it classifies each bounding box. The VGG-16 network \cite{Simonyan14c} was used to classify each bounding box.

\subsection{Multi-Person Tracking}
\label{sec:multi_person_tracking}

After detecting each person, a method is needed to track them in the scene. The main reasons why we need to track the detection is because it is necessary to remove false detections and continue tracking a non detected person from the object detection algorithm. 

We can assume that people are walking linearly in the scene to use a Kalman filter \cite{kalman1960new}. However since there will be at least two people per image, a multi-target tracking is needed. 

An approach which uses multiple Kalman filters and the Hungarian Algorithm \cite{kuo2010multi} was used to assign and unassign detected tracks.

\subsection{Human Pose Estimation}
\label{sec:human_pose_estimation}

To retrieve the human pose estimation, the work developed by Wei, \textit{et at.} "Convolutional Pose Machines" \cite{wei2016convolutional} was used. This paper presents a series of fully convolutional neural networks, each $n + 1$ stage depends in the output of the previous stage. It uses an idea similar to Tompson \textit{et al.} 2015 \cite{tompson2015efficient} giving the input image with different sizes. But this model creates a first network to get an initial prediction and use this information with the input image with different size on a second network, repeating it 12 times with different image sizes to refine the results. During training the gradients are propagated all over the network, however it has a loss layer after it stage to avoid the vanishing gradient problem \cite{Hochreiter:1998:VGP:353515.355233}. This work achieved  90.5\% of accuracy PCKh at 0.2. 

\subsection{Feature Extraction}
\label{sec:feature_extraction}

After computing the human pose estimation we can try to obtain meaningful information from it to be used as features. We need to find how to extract useful information from the two persons found in the scene.

Before extracting the features, it is needed to normalize the joint position in values between 0 and 1. It is done just diving the x value by the width and the y value by the height of the image. It was computed 6 different types of features.

\subsubsection{Joint Position}

The first, and maybe the most intuitive feature is the actual raw joint position from each person (Figure \ref{fig:joint_position}), which captures where each part of the human body is. Since in the dataset used the ankles were occluded, those joints were removed, so 12 joints were used.

\subsubsection{Distance related Joints}

The distance between related joints, means to capture the relation between right wrist, left wrist, left shoulder, right shoulder, for example. For interactions like shaking hands, it captures how those joints are related between each frame. The equation \ref{eq:drj} shows distance between related joints formula, where $P_{1}$ and $P_{2}$ contains the joint locations and $j$ is the joint index. It is possible to have a better visualization how it is computed in the Figure \ref{fig:drj}.

\begin{equation}\label{eq:drj}
F(j) = ||P_{1}(j) - P_{2}(j)||
\end{equation}

\subsubsection{Distance from one joint}

The distance from one joint captures the relation between a fixed joint from one person and all other joints in the other person. It means that it will capture the relative distance between one person and the other based on their joints. This feature is important when the distance between the two persons is an import information. In the Equation \ref{eq:doj} it is possible to see how it is calculated. In this Equation, $j_1$ and $j_2$ are the joint indexes where $j_1$ is always the head index and $j_2$ is the index for all the other parts. $P_{1}$ and $P_{2}$ contain the joint locations. In the Figure \ref{fig:doj} it is possible to have a graphical visualization how it is computed.

\begin{equation}\label{eq:doj}
F(j_1,j_2) = ||P_{1}(j_1) - P_{2}(j_2)||
\end{equation}

\subsubsection{Joint angles}

The other feature captured is the angle between the connected joints. This feature can capture the information about the joints, removing the joint location itself. It means that the person can make the interaction in any part of the image and it will not influence for this feature. The Equation \ref{eq:aj} shows how it is computed. In this Equation, $j_1$ and $j_2$ are the joint indexes where they represent connected joints, like right elbow and right wrist. $P_x$ and $P_y$ have the information about the $x$ and $y$ positions respectively. The Figure \ref{fig:aj} shows a graphical visualization how it is computed.

\begin{equation}\label{eq:aj}
F(j_1,j_2) = tan^{-1}\Bigg(\frac{P_y(j_1)-P_y(j_2)}{P_x(j_1)-P_x(j_2)}\Bigg)
\end{equation}

\subsubsection{Absolute difference}

The absolute difference features calculates the difference between related joints. This feature measures the distance in one dimension. In the Equation \ref{eq:ad} it is possible to see how it is calculated. $P$ contains the joint locations and $j$ is the joint index. 

\begin{equation}\label{eq:ad}
F(j) = |P_1(j) - P_2(j)|
\end{equation}

\subsubsection{Velocity}

\begin{figure}[h!]
 \centering
\begin{subfigure}{.22\textwidth}
  \includegraphics[width=1\linewidth]{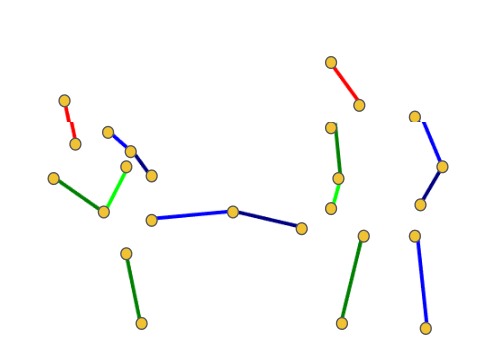}
  \caption{XY}
  \label{fig:joint_position}
\end{subfigure}%
\centering
\begin{subfigure}{.22\textwidth}
  \includegraphics[width=1\linewidth]{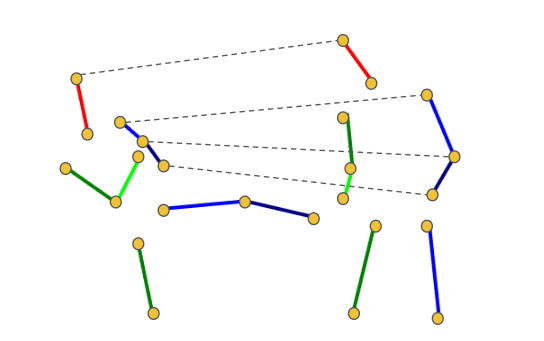}
  \caption{DRJ and AD}
  \label{fig:drj}
\end{subfigure}
 \centering
\begin{subfigure}{.22\textwidth}
  \includegraphics[width=1\linewidth]{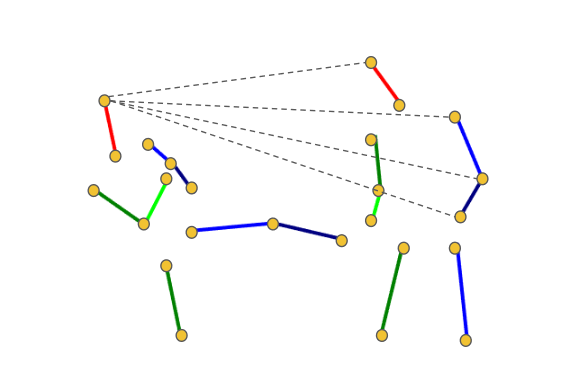}
  \caption{DOJ}
  \label{fig:doj}
\end{subfigure}

  \centering
\begin{subfigure}{.22\textwidth}
  \includegraphics[width=1\linewidth]{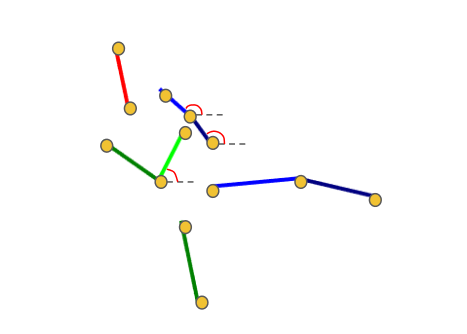}
  \caption{JA}
  \label{fig:aj}
\end{subfigure}
\centering
\begin{subfigure}{.22\textwidth}
  \centering
  \includegraphics[width=1\linewidth]{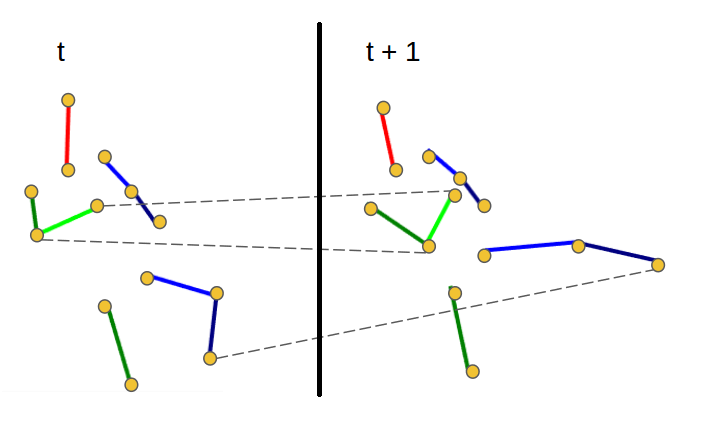}
  \caption{VEL}
  \label{fig:vel}
\end{subfigure}
\caption{Features extracted}
\label{fig:features}

\end{figure}

This feature gets the velocity of each joint between the frames. It captures the speed which can help to distinguish similar features where the velocity can be different. The equation \ref{eq:velocity} shows how it is computed. $P$ contains the information about the joint locations, $j$ is the joint index and $t_1$ and $t_2$ are the time indexes. In the Figure \ref{fig:vel} \footnote{The ankle velocity is just shown in this image for visualization purposes, the ankles were not used to compute the features} it is possible to visualize graphically how it is computed.

\begin{equation}\label{eq:velocity}
F(j,t_1,t_2) = P(j,t_1) - P(j,t_2)
\end{equation}

\subsection{Dataset}

The dataset used was the Two-person Interaction Dataset \cite{yun2012two}. It contains 282 samples in 21 sets of videos in 8 different types of interactions: approaching, departing, punching, kicking, hugging, pushing, shaking hands and exchanging an object. It was filmed with a static camera and always 2 persons per scene, and it also contained annotations of the human pose estimation in 3D which was not used.

\subsection{Interaction Classification}
\label{sec:interaction_classification}

To recognize the interaction the Support Vector Machine (SVM) \cite{cortes1995support} was used. The SVM has really interesting properties as a classifier. It is a linear classifier which has some support vectors to maximize the margin between the features. This classifier optimizes the margin between two classes creating a better generalization of the classification. The SVM used is the C-SVC, it uses a soft-margin approach, which allows some data inside the margin , the $C$ is a parameter that controls it. This important feature helps to create a better generalization in the classification, also helps to avoid outliers and overfitting. The kernel trick is also a important property of the SVM for non-linear classification, the kernel trick changes the basis of the features to a higher dimension where it can be linear separable.

\subsection{Implementation Details}
\label{sec:implementation_details}

All the development was made using MATLAB using a Ubuntu 14.04 Operational System. The computer used was a Intel(R) Xeon(R) CPU E5-2660 v3 @ 2.60GHz with 4 GPU's GeForce GTX TITAN. 

The implementation was done following each step of the methodology. It was done as follows:

\begin{itemize}

\item To detect the persons in each frame of the dataset, using the \textit{Faster RCNN} \cite{ren2015faster} took on average 0.1 seconds per frame.

\item Tracking multiple person was also done in the whole dataset took on average $0.8 ms$ per frame. A MATLAB implementation with default parameters was used for this task.

\item To estimate the human pose, it took 10 seconds per frame on average. This was the slowest part of the methodology. Caffe was used in this part \cite{jia2014caffe}.

\item To classify the interactions, it took $28 ms$ to train and $15 ms$ to test. LibSVM \cite{libsvm} in MATLAB was used in this task. The SVM parameters were C-SVC, Radial basis, $\gamma = 0.0625$ and $c = 8$.

\end{itemize}

\section{Results}

 After extracting various kinds of features, the different influence of each feature in the results will be studied. There will be two types of evaluation. The first is per frame evaluation which couple of consecutive frames will be evaluated. The second approach is the whole sequence evaluation in which the complete video as a whole will be classified. For the whole sequence evaluation it uses anchor frames equally spaced which all the samples will have the same number of frames to evaluated.
 
 Using the two-person dataset \cite{yun2012two} as reference, the influence of the number of frames to be used was calculated, it means, how many frames are necessary for a good evaluation. This dataset contains 21 sets, and it was created a 5-fold cross-validation using random sets to compute the accuracy, same approach used by \cite{yun2012two}. In this case just normalized raw joint positions from each person were used as features. Using the per frame evaluation with 9 frames it achieved 80.67\% of accuracy, it is not the best, with 20 frames it achieved 84.98\%. It was decided to use 9 frames, since the idea of this approach is to check how many consecutive frames is needed to achieve good results that could be useful in a real application. For the whole sequence evaluation approach, the best results were with 13 frames. It achieved 87.56\% of average accuracy of all interactions. 

The influence of each feature during the classification process was compared for each type of evaluation. The Table \ref{tab:acc_whole_sequence} shows all the combinations created and its result. XY means the raw $x$ and $y$ positions from the joints, DRJ means Distance from Related Joints, DOJ means Distance from One Joint, JA means Joint Angles, AD means Absolute difference, VEL means velocity.

\begin{table}[h!]
\centering
\caption{Accuracy results with different types of features}
\label{tab:acc_whole_sequence}
\begin{tabular}{l|l|l}
Features                  & \parbox[t]{1.2cm}{Accuracy\\  per frame}&\parbox[t]{1.2cm}{Accuracy\\ whole sequence}  \\ \hline
XY                        &80.67\% &\textbf{87.56\%}   \\
DRJ                       &74.57\% &82.67\%   \\
DOJ                       &78.98\% &84.73\%  \\
JA                        &64.83\% &69.22\%  \\
AD                        &79.20\% &84.00\%  \\
VEL                       &37.42\% &56.58\%  \\
XY + DRJ                  &81.16\% &85.55\%  \\
XY + DRJ + DOJ            &\textbf{81.75\%} &85.50\%  \\
XY + AD                   &81.38\% &85.49\%  \\
XY + DRJ + DOJ + AD       &81.61\% &83.32\%  \\
XY + DRJ + DOJ + VEL      &81.22\% &85.13\%  \\
All features              &71.77\% &81.68\%
\end{tabular}
\end{table}

\begin{figure*}[t]
\begin{subfigure}{0.5\textwidth}
  \centering
      \includegraphics[width=1\textwidth]{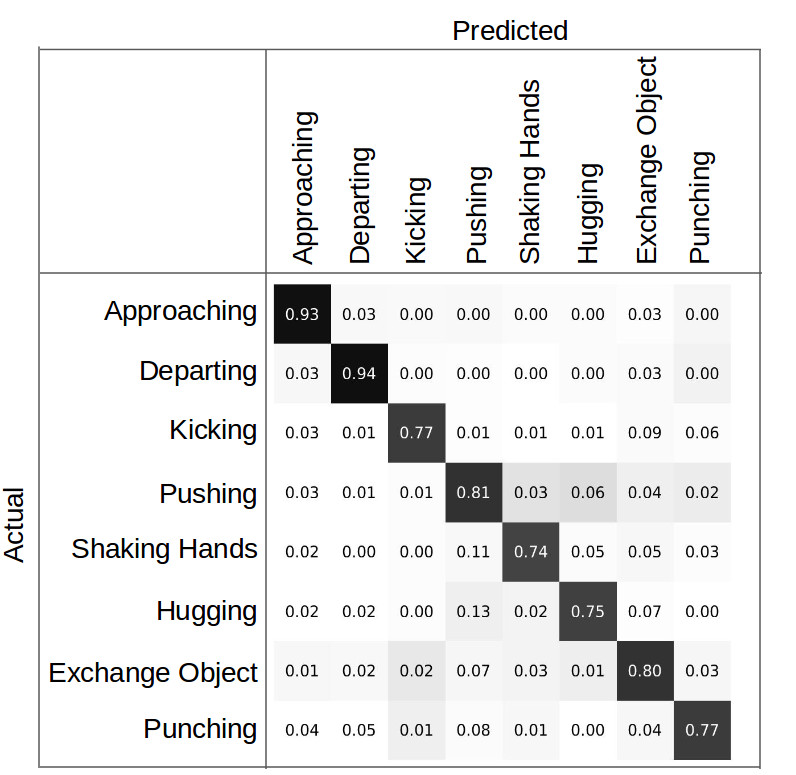}
  \caption{Confusion matrix for the per frame evaluation evaluation}\label{fig:per_frame_conf_matrix}
\end{subfigure}
\begin{subfigure}{0.5\textwidth}
  \centering
      \includegraphics[width=1\textwidth]{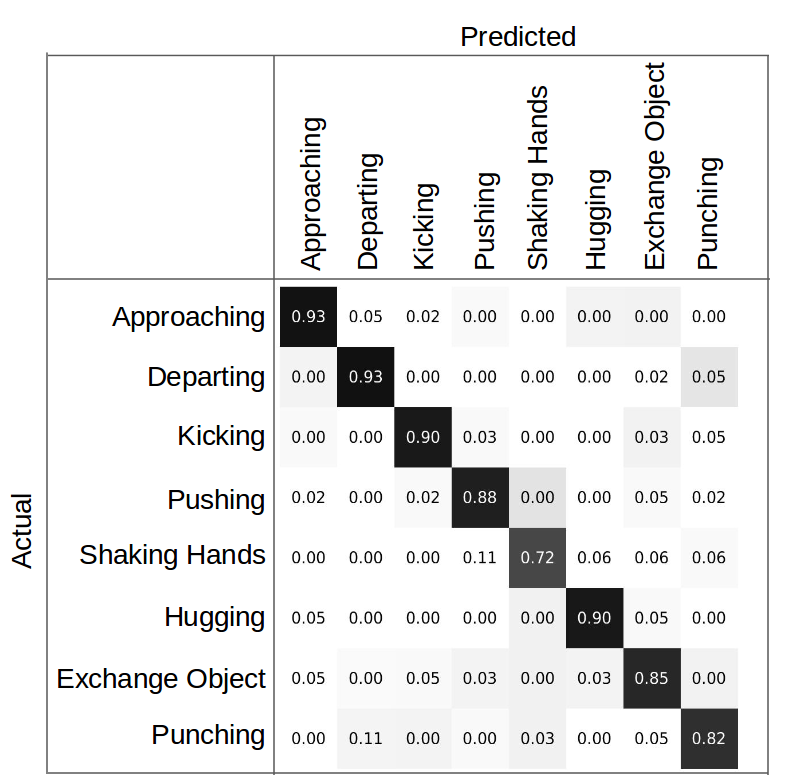}
  \caption{Confusion matrix for the whole sequence evaluation}\label{fig:conf_matrix_whole_sequence}
\end{subfigure}

\caption{Confusion matrix for the per frame evaluation (a) and whole sequence evaluation (b)}
\label{}
\end{figure*}

For the per frame evaluation, the raw $x$ and $y$ positions already show good results, all the other features depend on the $x$ and $y$ positions. The JA was the feature with lowest results, since our human pose estimation gives a 2D coordinate, the angle can give different values depending on how the person is "rotated" to the camera.

The DRJ and DOJ have some influence, however they cannot provide full information about the scene. Using XY, DCJ and DOJ provides the best results since this set of features can better capture meaningful information about the scene achieving 81.38 \%.

After achieving the best results using XY, DCJ and DOJ features, we can have a better look at the results looking at the confusion matrix (Figure \ref{fig:per_frame_conf_matrix}).

For the whole sequence evaluation the raw $x$ and $y$ positions can already get the best results. The other features did not help to get satisfactory results. Since the whole sequence evaluation gets spaced frames, the features computed loses a bit of the temporal meaning. All the other features depend on the $x$ and $y$ positions which explains why most of the results were really similar. 

The results from whole sequence evaluation were better than per frames evaluation because the whole sequence gets the complete information about the video, and the per frame only some part of the sequence is used which can be similar to other types of interaction.

The best results was got by only using the $x$ and $y$ position achieving 87.56\% of average accuracy between all interactions. In the Figure \ref{fig:conf_matrix_whole_sequence} it is possible to see the confusion matrix.

In comparison with other results Yun \textit{et at} \cite{yun2012two} who use the same dataset got 80.30\% of accuracy getting 3 frames to evaluate and 91.1\% using the whole sequence (Table \ref{tab:comparing_results}), however they uses the depth information to get the pose estimation. The method developed for this dissertation got 3.54\% of accuracy lower than the one developed by Yun, \textit{et al}, however my method only used RGB information. It outperformed using the per frame evaluation, however the method developed uses 9 frames, the one developed by Yun, \textit{et al} uses 3 frames.  It was computed the results using just 3 frames with the method developed dissertation and it achieved 76.14\% of accuracy, 3.16\% less than the method developed by Yun, \textit{et al}. This is due to depth information extracting better pose information thus getting better results. Comparing with the method developed by Hu, \textit{et al}, it outperformed in both evaluations. The Hidden Markov Model used to classify the interaction  is sensitive to noise and the skeleton captured by the kinect contains some noise, SVM is robust to some noise with the soft-margin approach.

\begin{table}[h!]
\centering
\caption{Results comparison}
\label{tab:comparing_results}
\begin{tabular}{l|ll}
                                    Method    & per frame & whole frame \\ \hline
Yun, \textit{et at} \cite{yun2012two}     & 80.30\%   & \textbf{91.10\%}     \\
Hu \textit{et at,} \cite{hu2013efficient} &   76.1\%        & 83.33\%          \\
Zhu, \textit{et at} \cite{zhu2015co}                                     &  \textbf{90.41\%}     &  -           \\
\textbf{Our Method} &  \textbf{81.75\%}   & \textbf{87.56\%}    
\end{tabular}
\end{table}


\section{CONCLUSIONS AND FUTURE WORKS}

\subsection{Conclusions}
This dissertation showed that using the deep convolutional neural networks to estimate the human pose can give great results to recognize human interaction using RGB camera. The method developed in this dissertation achieved 87.56\% of accuracy which is only 3.54\% worse than in the method developed by Yun \textit{et al} \cite{yun2012two} who use a depth camera to capture the human pose estimation. There are several steps which could be improved, but in general, the methodology was really intuitive and could achieve a result close to a method that uses a depth camera. The main problem with this method is the overlapping between two persons in some interactions which make it difficult to track and to estimate the pose. Also, the computation time is really expensive since it takes on average 10 seconds per frame to compute the pose estimation. The main contribution of this dissertation is that the recent development of human pose estimation could achieve close results to a method developed using a depth camera. The methodology presented in this document shows that retrieving the human pose from an RGB camera to recognize the interaction between two persons can be as effective as depth cameras.

\subsection{Future Works}

One approach which could achieve better results is to track each joint from the human body. Since videos are being used, it is possible to get the temporal information to increase the human pose estimation results. 

Many other types of features can be computed from the human joints. Yun \textit{et al}, describe 6 features from human joints which they captured to classify the interaction.

For the classification part, other classifiers can be used, such as Random Forest, or AdaBoost . However since a video is a sequence,  Hidden Markov Models and Long-Short Term Memory (LSTM) Recurrent Neural Networks are good classifiers for sequential data. Hu, \textit{et al} \cite{hu2013efficient} used HMM without successful results, however Zhu, \textit{et al} \cite{zhu2015co} achieved really good results using the LSTM showing that it can be a satisfactory approach to be used in future works.

\bibliographystyle{ieeetr}
\bibliography{refs}

\addcontentsline{toc}{chapter}
         {\protect\numberline{Bibliography\hspace{-96pt}}}

\end{document}